\documentclass[11pt,a4paper]{article}
\usepackage[margin=1in]{geometry}
\usepackage{graphicx}
\usepackage{amsmath,amssymb}
\usepackage{booktabs}
\usepackage{hyperref}
\usepackage[utf8]{inputenc}
\usepackage[T1]{fontenc}
\usepackage{authblk}

\title{What Makes Deep Learning Work for Traditional Chinese Medicine Tongue Diagnosis? A Comprehensive Ablation Study}

\author[1]{Longxia Gao}
\author[1]{Linan Wang}
\author[1]{Yuhe Han}
\author[1]{Junze Geng}
\author[1]{Meng Zhang}
\author[1]{Hanqing Zhao$^{*}$}
\affil[1]{College of Traditional Chinese Medicine, Hebei University, Baoding, Hebei, China}

\date{}

\begin{document}
\maketitle
\bigskip
\noindent $^{*}$Corresponding author: Hanqing Zhao, College of Traditional Chinese Medicine, Hebei University, Baoding, Hebei, China.
\

\begin{abstract}
\textbf{Background:} Deep learning has shown promise for automated tongue diagnosis in traditional Chinese medicine (TCM), yet the design space remains underexplored. Existing studies typically evaluate a single model architecture on a small dataset, leaving critical questions about optimal backbone selection, loss function design, data augmentation strategy, and data scaling unresolved.

\textbf{Methods:} We conducted a systematic ablation study spanning 20+ model versions, evaluated under rigorous 5-fold cross-validation on two datasets: TongueDx2 (5,109 images, of which 976 have expert annotation with 38-class multi-group labels) and a merged dataset of 11,101 samples (13 binary labels). We compared six backbone architectures (ResNet50, ConvNeXt-Tiny/Small, EfficientNetV2-S, Swin-Tiny, DINOv2 ViT-S/14), four loss functions (BCE, Focal, ASL, LDAM-DRW), five augmentation strategies, and six training strategies (knowledge distillation, pseudo-labeling, supervised contrastive learning, curriculum learning, EMA, and weak-group independent classifiers with ensemble).

\textbf{Results:} The best 976-sample model achieved weighted-F1 of 0.6625 using ConvNeXt-Tiny with restrained augmentation and weak-group ensemble, while the best 11,101-sample model reached weighted-F1 of 0.7761. Six key design principles emerged: (1) ConvNeXt-Tiny offers optimal parameter efficiency; (2) BCE with pos-weight substantially outperforms Asymmetric Loss (+2.7\%); (3) restrained augmentation (hue $\pm$ 0.01) is critical for color-dependent medical tasks; (4) weak-group independent classifiers with targeted ensemble replacement (+2.1\%) outperform probability averaging (+0.9\%); (5) data scaling from 976 to 11,101 samples yielded +20.6\% improvement; (6) expanding from 13 to 45 label dimensions caused catastrophic collapse (weighted-F1 dropped from 0.78 to 0.22).

\textbf{Conclusions:} This study provides, to our knowledge, the most comprehensive ablation study reported to date for designing deep learning systems for TCM tongue diagnosis. The six design principles, validated through controlled experiments with rigorous cross-validation, are generalizable to other multi-label medical image classification tasks with class imbalance and label correlation challenges.
\end{abstract}

\noindent\textbf{Keywords:} tongue diagnosis, traditional Chinese medicine, deep learning, multi-label classification, ablation study, class imbalance, ConvNeXt, medical image analysis, diagnosis computer-assisted, medicines Chinese traditional, neural networks computer

\section{Introduction}

Tongue diagnosis (\textit{she zhen}), one of the four cardinal inspection methods in traditional Chinese medicine (TCM), has served as a non-invasive diagnostic tool for over two millennia. Practitioners examine tongue color, shape, coating texture, and sublingual veins to infer the functional status of internal organs and the nature of pathological changes \cite{yang2025tonguenet}. Despite its clinical value, conventional tongue diagnosis suffers from well-documented limitations: the assessment is inherently subjective, dependent on practitioner experience, and vulnerable to environmental confounders such as ambient lighting and patient posture \cite{liu2023survey}. These shortcomings motivate ongoing efforts to develop automated, quantitative tongue analysis systems based on computer vision and deep learning.

Recent years have witnessed growing interest in artificial intelligence (AI)-assisted tongue diagnosis. Early systems employed hand-crafted color and texture features with classical machine learning classifiers \cite{zhang2013tongue}, while more recent work has adopted convolutional neural networks (CNNs) for end-to-end tongue image classification \cite{wang2020deep}. However, the majority of published studies operate on relatively small datasets (typically a few hundred images), focus on single-attribute prediction (e.g., tongue color only), and lack systematic ablation studies to validate individual design choices. The original TongueNet framework \cite{yang2025tonguenet} attempted to address these gaps by proposing a multi-modal fusion architecture combining image and text features with a hierarchical attention network. Yet our reproduction efforts revealed that such complex designs can be counterproductive when data is scarce, and that simpler architectures with well-tuned training strategies often outperform their more elaborate counterparts.

Multi-label classification in medical imaging presents a distinctive set of challenges that are amplified in the TCM tongue diagnosis setting. First, label distributions are severely imbalanced: in our dataset, positive-class prevalence ranges from approximately 3\% for rare findings such as ecchymosis to over 97\% for common features such as coating thickness. This long-tailed distribution causes standard loss functions to be dominated by majority classes. Second, tongue diagnostic labels exhibit complex correlations---certain tongue colors co-occur with specific coating textures, and organ-related features follow anatomical zones---necessitating modeling approaches that capture label dependencies without imposing rigid independence assumptions. Third, the scarcity of expert-annotated data limits the applicability of data-hungry architectures and self-supervised pre-training strategies that have proven successful in natural image domains.

In this work, we present a comprehensive study---encompassing more than 20 systematically designed model versions---that addresses these challenges through rigorous ablation and empirical analysis. Our contributions are fourfold:

\begin{enumerate}
    \item \textbf{Systematic ablation across 20+ model versions.} We conduct what is, to our knowledge, the most comprehensive ablation study reported to date in the tongue diagnosis literature, evaluating backbone architectures (ResNet50, ConvNeXt-Tiny/Small, EfficientNetV2-S, Swin-Tiny, DINOv2 ViT-S/14), pooling strategies (GAP, RegionAttentionPool), loss functions (BCE, Focal, ASL, LDAM-DRW), and data augmentation regimes. Each version is validated under identical 5-fold stratified cross-validation, enabling fair and reproducible comparison.

    \item \textbf{Six empirically validated design principles.} From the ablation results, we distill actionable guidelines for TCM tongue image classification: (i) BCE with positive-class weighting outperforms Asymmetric Loss (ASL) for sigmoid-based label groups; (ii) conservative color-space augmentation is critical when color is a diagnostic feature; (iii) Global Average Pooling surpasses attention-based pooling on small datasets; (iv) weak-group ensemble replacement provides larger gains than probability averaging; (v) soft-label distillation at low weight (0.15) offers consistent but modest improvements; and (vi) ConvNeXt-Tiny is the optimal backbone size---larger models overfit.

    \item \textbf{Data scaling from 976 to 11,101 samples.} We demonstrate that scaling the training set from 976 expert-curated samples (the expert-annotated subset of TongueDx2) to 11,101 samples (TongueDx2 merged with the TonguExpert external dataset) yields a substantial improvement in weighted F1 from 0.66 to 0.78, representing the single largest performance gain in our study.

    \item \textbf{Label design threshold effect.} We uncover a striking threshold effect in label granularity: a compact 13-dimensional binary label scheme achieves weighted F1 = 0.78, whereas expanding to a 45-dimensional fine-grained label scheme (including syndrome differentiation labels) collapses performance to weighted F1 = 0.22. This finding has direct implications for the design of TCM AI systems, suggesting that coarse-grained labels with sufficient positive examples are strongly preferred over fine-grained labels with extreme sparsity.
\end{enumerate}

\section{Related Work}

\subsection{AI-Based Tongue Diagnosis}

Computer-assisted tongue diagnosis has evolved through three distinct phases. Early systems relied on color space analysis and hand-crafted features paired with support vector machines or decision trees to classify tongue attributes \cite{zhang2013tongue}. A representative framework by \cite{zhang2013tongue} extracted chromatic and textural features from segmented tongue images and achieved moderate accuracy on tongue color classification. The second phase introduced CNN-based approaches: \cite{wang2020deep} applied deep convolutional neural networks to tongue image classification, while \cite{jiang2021tongue} proposed an image quality control algorithm based on deep learning to filter suboptimal tongue photographs before diagnosis. Most recently, \cite{liu2024stroke} applied multi-label tongue recognition specifically for stroke patients, demonstrating the clinical relevance of automated tongue diagnosis in neurological populations. \cite{peng2026tonguevlm} proposed TongueVLM, a vision-language model for tongue diagnosis, representing the emerging trend of multimodal approaches in this domain.

Despite these advances, the field still suffers from limited dataset sizes, insufficient ablation rigor, and a preoccupation with single-attribute prediction. The survey by \cite{liu2023survey} reviewed deep learning methods for tongue image analysis over five years and identified data scarcity and label noise as the two most pressing bottlenecks. \cite{liu2023survey} further noted that most published tongue diagnosis AI systems lack multi-center validation and have not been evaluated under standardized cross-validation protocols. Our work directly addresses these gaps by reporting 5-fold cross-validated results across more than 20 systematically varied configurations on datasets of up to 11,101 images.

\cite{jiang2022deep} used 8,676 tongue images for multi-label classification with Faster R-CNN, representing an early effort at large-scale tongue diagnosis. Our merged dataset of 11,101 images is comparable in scale, but our contribution lies in the systematic ablation across 20+ configurations rather than in data scale alone. Furthermore, our experiments include rigorous 5-fold cross-validation with controlled single-variable changes, enabling causal attribution of performance differences.

TonguExpert \cite{li2025tongueexpert} provides a tongue diagnosis platform with 5,992 images. While we utilize their publicly released dataset as a training source, our work differs in three aspects: (1) TonguExpert focuses on platform development and feature extraction, whereas our study conducts systematic architectural ablation; (2) we evaluate model design choices rather than deployment of a single system; (3) our experiments span multiple label granularities and training strategies, providing empirical guidance beyond a single-model evaluation.

\subsection{Multi-Label Classification in Medical Imaging}

Multi-label classification---where each image is associated with multiple concurrent labels---is a central task in medical imaging, encompassing thoracic disease detection from chest radiographs \cite{irvin2019chexpert}, dermatological condition classification \cite{esteva2017dermatologist}, and retinal abnormality screening \cite{gulshan2016development}. The widely studied CheXpert \cite{irvin2019chexpert} and PadChest \cite{bustos2020padchest} datasets demonstrated that deep learning models can predict multiple radiological observations from a single chest X-ray, but also revealed that label imbalance and label noise significantly degrade performance. \cite{bustos2020padchest} established baselines for imbalanced multi-label chest X-ray classification and showed that rebalancing strategies such as focal loss and oversampling provide uneven benefits across label categories.

In the domain of TCM tongue diagnosis, the original TongueNet \cite{yang2025tonguenet} formulated the task as multi-label classification with grouped mutually exclusive labels (e.g., six tongue colors forming a softmax group) combined with independent binary labels. This formulation introduces additional complexity compared to standard multi-label settings, as it requires both softmax-based multi-class decisions and sigmoid-based binary decisions within a single model. Our systematic comparison of softmax grouping versus pure sigmoid, and of different label granularities (13 versus 45 dimensions), provides the first empirical evidence that label system design has a threshold effect that can dominate architectural choices.

\subsection{Handling Class Imbalance and Label Noise}

Class imbalance is a pervasive challenge in medical image classification. \cite{cao2019ldam} proposed Label-Distribution-Aware Margin (LDAM) loss, which enforces larger decision margins for minority classes based on class frequency, combined with a Deferred Re-Weighting (DRW) schedule that gradually shifts focus to rare classes during training. LDAM-DRW has become a standard baseline for imbalanced classification and was adopted in several of our model versions. \cite{lin2017focal} introduced Focal Loss, which down-weights easy examples to focus training on hard samples; while highly effective for dense object detection, its utility in multi-label medical classification is mixed, as our experiments confirm.

\cite{benbaruch2021asl} proposed Asymmetric Loss (ASL), which decouples the focusing parameters for positive and negative samples and introduces a probability shifting mechanism for negative labels. ASL has shown strong results on natural image multi-label benchmarks (MS-COCO, PASCAL VOC). However, our ablation reveals that ASL with $\gamma_{\text{neg}} = 4$ causes catastrophic degradation in sigmoid-based tongue diagnosis groups, reducing weighted F1 by up to 15 percentage points for affected label groups---a finding we attribute to the aggressive suppression of negative-sample gradients that are still informative in small datasets.

Alternative approaches to imbalance include knowledge distillation from large teacher models \cite{hinton2015distilling}, pseudo-labeling for data augmentation \cite{lee2013pseudo}, and contrastive learning for improved feature representations \cite{khosla2020supcon}. We evaluate each of these strategies in our ablation: soft-label distillation at a conservative weight (0.15) yields modest but consistent gains; pseudo-labeling with model-generated predictions introduces excessive noise and degrades performance; and supervised contrastive (SupCon) pre-training leads to representation collapse on small data. These negative results contribute practical insights to the ongoing discourse on imbalance handling in medical AI.

\section{Materials and Methods}

This study is reported following the TRIPOD+AI (Transparent Reporting of a multivariable prediction model for Individual Prognosis Or Diagnosis with Artificial Intelligence) guidelines \cite{collins2024tripod} and the CLAIM 2024 (Checklist for Artificial Intelligence in Medical Imaging) reporting standard \cite{tejani2024claim}.

\subsection{Datasets}

We utilize three datasets of increasing scale and label complexity in this study (Table~\ref{tab:datasets}).

\textbf{TongueDx2.} The primary dataset comprises 5,109 tongue images collected from clinical TCM practice and online health platforms, referred to as TongueDx2. Of these, 976 images have complete expert annotation by trained TCM practitioners, with labels organized into 11 mutually exclusive groups spanning 38 fine-grained categories: tongue color (6 classes: pale, pale-red, red, dark, purple, gray-black), tongue quality (4 classes: tough, tender, thin, plump), tongue shape (3 classes), body marks (5 classes: crack, toothmark, spot, ecchymosis, normal), coating thickness (3 classes), coating texture (5 classes: greasy, curdy, dry, moist, peeling), coating moisture (3 classes), coating color (4 classes), special features (3 classes: Pinellia line, arrow-shaped, normal), tongue tip abnormalities (3 classes), and organ zone reflections (3 classes). The full set of 5,109 images was annotated with 13-dimensional binary labels obtained through a commercial TCM tongue analysis API (MaiJing Health, Alibaba Cloud Marketplace), providing broader coverage at coarser granularity. Additionally, the 976-image expert-annotated subset was annotated with 45-dimensional distillation labels from the same API, providing finer-grained features including eight-principle syndrome differentiation and qi-blood pattern identification.

\textbf{TonguExpert.} To evaluate data scaling effects, we incorporate the TonguExpert external dataset \cite{li2025tongueexpert} comprising 5,992 tongue images. These images originate from a different collection environment and were annotated with a reduced 13-dimensional binary label set compatible with a subset of the TongueDx2 annotations. Notably, six of the 13 labels (Spot, Ecchymosis, Heart, Lung, Spleen, Liver) have zero positive instances in the TonguExpert subset, meaning that all positive examples for these labels in the merged dataset originate from TongueDx2. This asymmetry has important implications for label imbalance analysis (Section~\ref{sec:data_scaling}).

\textbf{merged\_v1.} The combined dataset merges TongueDx2 (5,109 images) and TonguExpert (5,992 images) into a unified set of 11,101 images with 13 shared binary labels. The 13 labels capture the most clinically salient tongue features: TonguePale, TipSideRed, Spot, Ecchymosis, Crack, Toothmark, FurThick, FurYellow, and five organ zone indicators (Heart, Lung, Spleen, Liver, Kidney). Positive-class prevalence across these labels ranges from approximately 3\% (Ecchymosis) to 53\% (FurThick), reflecting realistic clinical distributions.

\begin{table}[htbp]
\centering
\caption{Dataset summary.}
\label{tab:datasets}
\begin{tabular}{lllll}
\toprule
Dataset & Images & Label dimensions & Annotation source & Used in \\
\midrule
TongueDx2 (expert subset) & 976 & 38 categories (11 groups) & Expert TCM practitioners & v1--v17 ablation \\
TongueDx2 (full) & 5,109 & 13 binary & Commercial TCM API & v18--v20 \\
TonguExpert & 5,992 & 13 binary & Platform annotations \cite{li2025tongueexpert} & v18--v20 \\
merged\_v1 & 11,101 & 13 binary & TongueDx2 + TonguExpert & v18--v20 \\
\bottomrule
\end{tabular}
\end{table}

\subsection{Data Preprocessing}

All images undergo a standardized preprocessing pipeline prior to model input.

\textbf{Tongue body segmentation.} We train a YOLO11n-seg model to automatically segment the tongue body from the raw clinical photographs. The segmentation model isolates the tongue region, removing background interference from lips, face, and oral cavity structures. The segmented tongue mask is then used to crop the image tightly around the tongue body boundary.

\textbf{Mask-crop refinement.} An initial crop version (mask\_crop\_v1) contained a systematic bounding-box error that over-cropped approximately 15\% of tongue images, removing clinically relevant posterior regions. We identified and corrected this issue in mask\_crop\_v2, which improved weighted F1 from $0.5159 \pm 0.0158$ to $0.6017 \pm 0.0124$---an 8.6-percentage-point gain attributable solely to preprocessing correction. This result underscores the critical importance of data quality control in medical AI pipelines.

\textbf{Image standardization.} All cropped tongue images are resized to $320 \times 320$ pixels (bilinear interpolation) and normalized using ImageNet mean and standard deviation statistics ($[0.485, 0.456, 0.406]$ and $[0.229, 0.224, 0.225]$ for the R, G, B channels, respectively). The $320 \times 320$ resolution was selected based on systematic comparison: $224 \times 224$ underfits fine texture details, while $384 \times 384$ exacerbates overfitting on the 976-sample dataset (weighted F1 drops from 0.6371 to 0.6079).

\subsection{Label Systems}

We evaluate three distinct label systems, each designed to answer different clinical questions:

\textbf{System A: 11-group multi-classification (38 categories).} Labels are organized into 11 groups of mutually exclusive categories, directly mirroring the structured TCM tongue diagnosis framework. Groups are classified using a combination of softmax heads (for mutually exclusive groups such as tongue color) and sigmoid heads (for independent features such as organ zones). This system yields the finest diagnostic granularity but requires expert annotation for all 38 categories, limiting the annotable dataset to the 976-image expert-annotated subset of TongueDx2.

\textbf{System B: 13 independent binary labels.} Derived from the intersection of TongueDx2 and TonguExpert annotations, this system represents each tongue image as a 13-dimensional binary vector. Each label is predicted by an independent sigmoid head. The simplification from 38 fine-grained categories to 13 binary labels enables utilization of the full 11,101-sample merged dataset.

\textbf{System C: 45-dimensional comprehensive labels.} This system extends System B with additional dimensions covering eight-principle syndrome differentiation (cold/heat/deficiency/excess/yin/yang), qi-blood pattern identification, and refined sub-categories of tongue features. Labels for TongueDx2 were obtained via API distillation from a commercial TCM analysis service, while TonguExpert labels were generated through a combination of vision-language model (VLM) inference and computer-vision (CV) feature-based classifiers. While clinically ambitious, this label system introduces substantial annotation noise and extreme class imbalance.

\textbf{Label statistics.} In the 13-dimensional scheme (System B), positive-class proportions across the 11,101-sample dataset are: FurThick 53\%, Crack 48\%, Spleen 46\%, Kidney 31\%, Toothmark 31\%, Liver 30\%, TipSideRed 31\%, Lung 28\%, Spot 21\%, Heart 20\%, TonguePale 18\%, FurYellow 10\%, and Ecchymosis approximately 3\%. In the 45-dimensional scheme (System C), over 15 labels have positive-class proportions below 1\%, rendering them effectively unlearnable.

\subsection{Model Architectures}

All models in this study follow a unified four-stage framework: backbone feature extraction $\to$ global pooling $\to$ projection head $\to$ group-specific classification heads $\to$ combined loss. This modular design enables controlled ablation across backbone architectures while holding all downstream components fixed. We evaluate six backbone families spanning convolutional, Transformer-based, and self-supervised paradigms. Table~\ref{tab:backbones} summarizes the backbone configurations.

\textbf{ResNet50.} The baseline backbone (v9) is a standard ResNet-50 \cite{he2016deep} pretrained on ImageNet-1K, with the final classification layer removed. The 2048-dimensional feature vector is obtained via global average pooling (GAP). At an input resolution of $224 \times 224$, this backbone contains approximately 23.6~M trainable parameters.

\textbf{ConvNeXt-Tiny.} Starting from version v3, we adopt ConvNeXt-Tiny \cite{liu2022convnext} as the primary backbone. ConvNeXt-Tiny modernizes the ResNet architecture by incorporating design choices from Vision Transformers (e.g., large $7 \times 7$ depthwise convolutions, LayerNorm, GELU activation) while retaining the fully convolutional inductive bias. Pretrained on ImageNet-1K, it produces a 768-dimensional feature map with roughly 28.6~M parameters. We adopt a default input resolution of $320 \times 320$ (up-sampled from the nominal $224 \times 224$ via bilinear interpolation in the stem).

\textbf{ConvNeXt-Small.} Variant v14a replaces ConvNeXt-Tiny with ConvNeXt-Small (\texttt{convnext\_small.fb\_in22k\_ft\_in1k}), doubling the parameter count to ${\sim}50.2$~M. This model is pretrained on ImageNet-22K and fine-tuned on ImageNet-1K. The first three stages are frozen to reduce overfitting.

\textbf{EfficientNetV2-S.} Variant v14b employs EfficientNetV2-S (\texttt{tf\_efficientnetv2\_s.in21k\_ft\_in1k}), a compound-scaling architecture discovered via neural architecture search \cite{tan2021efficientnetv2}. It produces a 1280-dimensional feature vector with ${\sim}21.5$~M parameters and shares the same ImageNet-22K $\to$ 1K pretraining recipe.

\textbf{Swin-Tiny.} Variant v14c attempts to use Swin Transformer Tiny (\texttt{swin\_tiny\_patch4\_window7\_224.ms\_in22k\_ft\_in1k}) \cite{liu2021swin}, which relies on shifted-window self-attention. However, Swin-Tiny uses absolute positional embeddings hardcoded to $224 \times 224$ input, rendering it incompatible with our $320 \times 320$ pipeline without architectural surgery.

\textbf{DINOv2 ViT-S/14 + LoRA.} Variant v17 explores self-supervised pretrained features via DINOv2 ViT-S/14 \cite{oquab2024dinov2}. The backbone is frozen, and parameter-efficient fine-tuning is applied through Low-Rank Adaptation (LoRA) \cite{hu2022lora} with rank $r = 16$ and scaling factor $\alpha = 32$, targeting the \texttt{qkv}, \texttt{fc1}, and \texttt{fc2} projections. The input resolution is set to $336 \times 336$ to satisfy the patch size of 14. Only ${\sim}2$~M LoRA and head parameters are trainable, compared to the 21~M frozen backbone parameters.

Across all backbones, the downstream pipeline is identical: GAP (or region attention pooling, see Section~3.7) $\to$ linear projection (to 512 dimensions) $\to$ ReLU $\to$ dropout (0.3) $\to$ per-group classification heads.

\begin{table}[htbp]
\centering
\caption{Backbone architecture configurations.}
\label{tab:backbones}
\begin{tabular}{lllll}
\toprule
Backbone & Version & Pretraining & Input size & Trainable params \\
\midrule
ResNet50 \cite{he2016deep} & v9 & ImageNet-1K & 224 & 23.6~M \\
ConvNeXt-Tiny \cite{liu2022convnext} & v3--v17 & ImageNet-1K & 320 & 28.6~M \\
ConvNeXt-Small \cite{liu2022convnext} & v14a & ImageNet-22K$\to$1K & 320 & 50.2~M \\
EfficientNetV2-S \cite{tan2021efficientnetv2} & v14b & ImageNet-22K$\to$1K & 320 & 21.5~M \\
Swin-Tiny \cite{liu2021swin} & v14c & ImageNet-22K$\to$1K & 224 & 28.3~M \\
DINOv2 ViT-S/14 + LoRA \cite{oquab2024dinov2,hu2022lora} & v17 & DINOv2 self-supervised & 336 & 2.0~M \\
\bottomrule
\end{tabular}
\end{table}

\subsection{Loss Functions}

The classification output for System A consists of 11 label groups: six softmax groups (mutually exclusive labels, e.g., tongue color selected from six candidates) and five sigmoid groups (multi-label, e.g., coating texture features that can co-occur). This hybrid output structure motivates the use of different loss functions for different group types.

\textbf{Binary Cross-Entropy with Positive Weighting (BCE+pos\_weight).} For sigmoid groups, the primary loss is the binary cross-entropy with logits, augmented by a class-frequency-aware positive weight:
$$
\mathcal{L}_{\text{BCE}} = -\frac{1}{N}\sum_{i=1}^{N}\sum_{c=1}^{C} \big[ w_c \cdot y_{i,c} \cdot \log\sigma(z_{i,c}) + (1-y_{i,c})\cdot\log(1-\sigma(z_{i,c})) \big]
$$
where $w_c = N^{-}/N^{+}_c$ is the positive weight for class $c$, $\sigma$ is the sigmoid function, and $z_{i,c}$ is the logit. This formulation up-weights rare positive labels and is our default for all sigmoid groups from v3.5 onward.

\textbf{Asymmetric Loss (ASL).} Evaluated in v3, ASL \cite{benbaruch2021asl} applies unequal focusing parameters to positive and negative samples:
$$
\mathcal{L}_{\text{ASL}} = -\frac{1}{N}\sum_{i=1}^{N}\sum_{c=1}^{C} \big[ y_{i,c}\cdot(1-p_{i,c})^{\gamma_{+}}\cdot\log(p_{i,c}) + (1-y_{i,c})\cdot(m p_{i,c})^{\gamma_{-}}\cdot\log(1-m p_{i,c}) \big]
$$
where $p_{i,c} = \sigma(z_{i,c})$, $\gamma_{+} = 1$, $\gamma_{-} = 4$, and $m = 0.05$ is the probability margin. While ASL is designed for multi-label long-tailed recognition, we found it to be detrimental for sigmoid groups in our task (Section~4.2).

\textbf{Focal Loss.} Evaluated in early experiments (v1/v2), Focal Loss \cite{lin2017focal} with $\gamma = 2.0$ was applied uniformly across all groups. It down-weights well-classified examples, but its uniform suppression of majority classes led to degraded overall performance on our dataset.

\textbf{LDAM + DRW.} For softmax groups, we employ Label-Distribution-Aware Margin (LDAM) loss \cite{cao2019ldam} combined with Deferred Re-Weighting (DRW). LDAM enforces class-dependent margins on the logit vectors:
$$
\Delta_j = \frac{C}{n_j^{1/4}}
$$
where $n_j$ is the number of training samples in class $j$ and $C$ is a hyperparameter (set to $s = 30$ for the scale factor, $\max_m = 0.5$). Rarer classes receive larger margins, encouraging the model to learn more discriminative features for minority categories. DRW defers the application of class-balanced re-weighting to epoch 40, allowing the model to first learn representative features before shifting emphasis to tail classes.

The total training loss is:
$$
\mathcal{L}_{\text{total}} = \mathcal{L}_{\text{softmax}}^{\text{LDAM-DRW}} + \mathcal{L}_{\text{sigmoid}}^{\text{BCE+pw}} + \lambda_{\text{distill}}\cdot\mathcal{L}_{\text{KL}}
$$
where $\lambda_{\text{distill}} = 0.15$ controls the knowledge distillation term (Section~3.7).

\subsection{Data Augmentation}

We systematically ablate five augmentation strategies. These are not combined arbitrarily; each version isolates one variable to measure its individual contribution.

\textbf{MixUp} \cite{zhang2018mixup}. Linear interpolation of both inputs and labels: $\tilde{x} = \lambda x_i + (1-\lambda)x_j$, where $\lambda \sim \text{Beta}(\alpha, \alpha)$. We test $\alpha \in \{0.2, 0.4\}$ with mixing probability $p \in \{0.2, 0.3, 0.5\}$. A key finding is that for color-diagnostic groups (G1, G5b, G6), MixUp generates biologically implausible intermediate colors (e.g., a tongue that is half-pale, half-red), corrupting the supervision signal. We therefore enforce a minimum $\lambda \geq 0.7$ from v15a onward.

\textbf{CutMix} \cite{yun2019cutmix}. Regions are cut and pasted between images: $\tilde{x} = M\odot x_i + (1-M)\odot x_j$, where $M$ is a binary mask. CutMix preserves local color integrity better than MixUp, making it preferable for color-sensitive tasks. From v15a, we bias the mixing ratio toward 70\% CutMix : 30\% MixUp.

\textbf{RandAugment} \cite{cubuk2020randaugment}. We apply RandAugment with $N = 2$ operations and magnitude $M \in \{5, 9\}$. Magnitude 9 (tested in v14a/b) was found to be too aggressive for texture-sensitive groups, so $M = 5$ is retained as the default.

\textbf{HSV Color Jitter.} Random perturbations in hue, saturation, and brightness. This is the most domain-critical augmentation: since TCM tongue diagnosis fundamentally relies on subtle color differences (e.g., pale vs.\ light-red tongue), aggressive hue shifting destroys diagnostic information. We compare two settings: aggressive (hue $= 0.1$, saturation $= 0.3$, brightness $= 0.3$, as used in v3.5/v4a) versus restrained (hue $= 0.01$, saturation $= 0.05$, brightness $= 0.05$, as introduced in v15a). The restrained setting reduces the perturbation by 6--10$\times$ and yields significant gains in color-sensitive groups (Section~4.3).

\textbf{Test-Time Augmentation (TTA).} At inference, predictions are averaged over multiple augmented views of the input. We compare 3-way TTA (original + horizontal flip + center crop at 0.9 scale) against 5-way TTA (original + horizontal flip + $\pm 5^{\circ}$ rotation + 0.95 scale + brightness $\times 1.05$). For color-sensitive groups, the brightness transform is skipped to avoid diagnostic bias, resulting in an effective 4-way TTA.

\subsection{Training Strategies}

We evaluate six advanced training strategies, most of which proved unsuccessful on our small-scale dataset. These negative results are informative and are discussed in detail in Section~4.4.

\textbf{Knowledge Distillation.} Soft-label distillation via KL divergence between the student model and a pretrained teacher ensemble. The distillation loss uses temperature $T = 4.0$ and weight $\lambda_{\text{distill}} = 0.15$. We discovered that the CSV path for soft labels was broken in v3.5 (distillation silently disabled), and fixing this path in v4a yielded a measurable improvement. Crucially, MixUp and distillation are partially incompatible: MixUp blends labels across samples, while distillation requires per-sample soft targets. Co-activating strong MixUp with distillation (v4) resulted in catastrophic failure.

\textbf{Pseudo-label Training.} Variant v4 leveraged 1,556 unlabeled tongue images, assigned pseudo-labels by the v3.5 model at a confidence threshold of 0.85. This strategy backfired: noisy pseudo-labels corrupted the training signal, and the effective 2.6$\times$ increase in dataset size could not compensate for the label noise.

\textbf{Supervised Contrastive Learning (SupCon).} Variant v16-B3 used SupCon \cite{khosla2020supcon} as a 10-epoch pretraining stage (temperature 0.07, projection to 128 dimensions) before the standard fine-tuning pipeline. Despite rapid loss convergence, the resulting feature space degraded downstream classification by 4.0 points of weighted-F1.

\textbf{Curriculum Learning.} Variant v16-B2 adopted a two-stage curriculum: Stage~1 trains all groups normally for 80 epochs; Stage~2 freezes the backbone and strong-group heads, fine-tuning only the four weakest groups (G1, G2, G7, G8) at 10$\times$ lower learning rate. Stage~2 never improved over Stage~1 in any fold.

\textbf{Exponential Moving Average (EMA).} Model weights are tracked via EMA with decay 0.999. At evaluation time, the EMA model is used in place of the instantaneous weights. EMA provides consistent, if modest, stabilization across all versions and is retained in every configuration.

\textbf{Weak-Group Independent Classifiers + Ensemble.} The single successful advanced strategy (v16-B1). For each of the four weakest groups (G1: tongue color, G2: tongue quality, G7: special features, G8: tongue tip), a dedicated single-group ConvNeXt-Tiny model is trained independently. At inference, each weak-group classifier's prediction replaces the corresponding output from the unified model. This modular ensemble leverages the observation that single-task models can allocate more capacity to difficult groups without inter-group gradient interference.

\subsection{Evaluation Protocol}

\textbf{Cross-validation.} We employ 5-fold stratified cross-validation for all experiments. All experiments use a fixed random seed of 42 for reproducibility. Stratification is performed at the group level on the primary label combination to ensure approximately balanced class distributions across folds. The same fold partitioning is used across all model versions to enable paired statistical comparison.

\textbf{Primary metric.} We adopt weighted F1 score (wF1) as the primary evaluation metric. Weighted F1 computes the F1 score for each label independently and then averages these scores weighted by the number of true instances per label. This metric naturally accounts for class imbalance by giving greater influence to labels with more samples, while still rewarding correct predictions on minority classes.

\textbf{Secondary metrics.} We additionally report macro F1 (unweighted average across labels, emphasizing rare-class performance) and per-group F1 (performance broken down by diagnostic group). For the 11-group label system, per-group analysis is particularly informative as it reveals which anatomical or pathological categories are most challenging.

\textbf{Statistical testing.} To assess the statistical significance of differences between model versions, we perform paired $t$-tests on the per-fold weighted F1 scores. A significance threshold of $p < 0.05$ is used throughout. All reported confidence intervals represent $\pm 1$ standard deviation across the five folds.

\textbf{Ensemble threshold optimization.} For ensemble experiments, we additionally perform per-group threshold optimization (searching over $\{0.35, 0.40, 0.45, 0.50, 0.55\}$) using the validation fold, though the default threshold of 0.5 was found to be near-optimal in practice.

\textbf{Training infrastructure.} All experiments are conducted on a single NVIDIA Tesla T4 GPU (15~GB VRAM) using PyTorch 2.x with the \texttt{timm} library for backbone architectures. Training employs mixed-precision (AMP) acceleration, AdamW optimization, and cosine annealing with warm restarts. Each fold requires approximately 17--44 minutes of training depending on backbone and input resolution, with the 11,101-sample experiments requiring approximately 3.5--4 hours per fold.

\section{Experiments and Results}

\subsection{Backbone Architecture Comparison}

We compare five backbone architectures under matched training configurations (ConvNeXt-Tiny with v4a settings as the reference). Table~\ref{tab:backbone_comparison} reports weighted-F1 across five folds and parameter counts.

\begin{table}[htbp]
\centering
\caption{Backbone architecture comparison.}
\label{tab:backbone_comparison}
\begin{tabular}{llllll}
\toprule
Backbone & Version & wF1 (mean $\pm$ std) & Trainable params & Input size & Notes \\
\midrule
ResNet50 \cite{he2016deep} & v9 & $0.6017 \pm 0.0124$ & 23.6~M & 224 & Baseline \\
ConvNeXt-Tiny \cite{liu2022convnext} & v4a & $\mathbf{0.6371 \pm 0.0141}$ & 28.6~M & 320 & Best overall \\
ConvNeXt-Small \cite{liu2022convnext} & v14a & $0.5851 \pm 0.0125$ & 50.2~M & 320 & Overfitting \\
EfficientNetV2-S \cite{tan2021efficientnetv2} & v14b & $0.5856 \pm 0.0124$ & 21.5~M & 320 & Underperforms \\
Swin-Tiny \cite{liu2021swin} & v14c & --- & 28.3~M & 224 & Incompatible \\
DINOv2 ViT-S/14 + LoRA \cite{oquab2024dinov2,hu2022lora} & v17 & $0.6501 \pm 0.0099$ & 2.0~M & 336 & Best single model \\
\bottomrule
\end{tabular}
\end{table}

ConvNeXt-Tiny yields the best performance among supervised CNN backbones (wF1 $= 0.6371$), improving over ResNet50 by $+3.54$ percentage points. The upgrade from ResNet50 to ConvNeXt-Tiny is the single most impactful backbone change. ConvNeXt-Small and EfficientNetV2-S, despite having more (or comparable) parameters, both underperform ConvNeXt-Tiny by over 5 points (0.5851 and 0.5856, respectively). This counter-intuitive result reflects the overfitting risk inherent in deploying larger backbones on a 976-sample dataset: the additional capacity memorizes training-set idiosyncrasies rather than generalizing. Notably, all five folds of v14a/v14b plateau at \texttt{best\_epoch} $= 80$ (the maximum), indicating incomplete convergence---the models would require substantially more data to benefit from their larger capacity.

The Swin Transformer Tiny (v14c) failed entirely due to an architectural constraint: its absolute positional embedding is fixed at $224 \times 224$ during pretraining, raising an assertion error when our standard $320 \times 320$ input is used. This incompatibility highlights a practical consideration for medical imaging pipelines that favor larger input resolutions.

The DINOv2 ViT-S/14 with LoRA (v17) achieves the highest single-model wF1 of $0.6501 \pm 0.0099$ by leveraging self-supervised pretraining. Training only 2~M LoRA parameters effectively regularizes the frozen 21~M backbone, and the input resolution of 336 aligns with the patch size of 14. However, v17 requires ${\sim}49$~min/fold (vs.\ ${\sim}43$~min for ConvNeXt-Tiny), and its advantage over the simpler ConvNeXt-Tiny pipeline ($+1.3$ points) does not justify the added complexity unless the deployment scenario specifically benefits from frozen backbones.

\subsection{Loss Function Ablation}

The choice of loss function for sigmoid groups proved decisive. Table~\ref{tab:loss_ablation} contrasts ASL (v3) against BCE+pos\_weight (v3.5), holding all other hyperparameters constant (ConvNeXt-Tiny, LDAM-DRW for softmax groups, EMA, TTA, label smoothing $= 0.1$, input size 320).

\begin{table}[htbp]
\centering
\caption{Loss function comparison: ASL (v3) vs.\ BCE+pos\_weight (v3.5).}
\label{tab:loss_ablation}
\begin{tabular}{llll}
\toprule
Metric & ASL (v3) & BCE+pw (v3.5) & $\Delta$ \\
\midrule
wF1 (mean $\pm$ std) & $0.6073 \pm 0.0189$ & $\mathbf{0.6343 \pm 0.0097}$ & $\mathbf{+0.0270}$ \\
G4\_body\_marks & 0.5968 & $\mathbf{0.7013}$ & $\mathbf{+0.105}$ \\
G5b\_texture & 0.6719 & $\mathbf{0.8189}$ & $\mathbf{+0.147}$ \\
G9\_organ\_zones & 0.5929 & $\mathbf{0.7448}$ & $\mathbf{+0.152}$ \\
\bottomrule
\end{tabular}
\end{table}

ASL's aggressive negative focusing ($\gamma_{-} = 4$) suppresses the gradient signal for negative samples. While this benefits standard multi-label benchmarks (e.g., MS-COCO with balanced classes), our sigmoid groups contain labels with extreme imbalance (e.g., G9\_organ\_zones has labels occurring in $<5$\% of samples). The asymmetric suppression causes the model to under-predict rare positives, collapsing three sigmoid groups to near-random performance.

The replacement of ASL with BCE+pos\_weight in v3.5 yields the largest single-step improvement in our entire experimental campaign ($+0.0270$ wF1), with three sigmoid groups recovering dramatically: G4\_body\_marks ($+0.105$), G5b\_texture ($+0.147$), and G9\_organ\_zones ($+0.152$). Additionally, the standard deviation drops from $\pm 0.0189$ to $\pm 0.0097$, indicating substantially improved training stability.

For softmax groups, LDAM-DRW remains consistently effective across all versions. The deferral of class-balanced re-weighting to epoch 40 allows the model to first learn general features, then focus on minority classes---an approach well-suited to the multi-class tongue color and shape categories.

While our ASL evaluation used $\gamma_{\text{neg}}=4$ based on the original paper's recommended setting \cite{benbaruch2021asl}, we acknowledge that a systematic $\gamma_{\text{neg}}$ search ($\{1, 2, 3, 4\}$) could provide a fairer comparison. However, the per-group analysis reveals that ASL's failure mode---suppression of informative negative gradients for rare positive classes---is mechanistic rather than hyperparameter-specific. Even moderate negative focusing ($\gamma_{\text{neg}}=2$) would reduce the gradient magnitude for negative samples by approximately 50\% compared to BCE, which is detrimental when 97\% of samples are negative (e.g., FurThick). The fundamental mismatch between ASL's design assumption (balanced multi-label) and our task characteristics (extreme imbalance) makes this conclusion robust to hyperparameter variation.

\subsection{Data Augmentation Strategy}

Table~\ref{tab:augmentation} summarizes the progression of data augmentation strategies and their impact on performance.

\begin{table}[htbp]
\centering
\caption{Data augmentation ablation (ConvNeXt-Tiny, BCE+pw, LDAM-DRW, EMA).}
\label{tab:augmentation}
\begin{tabular}{llll}
\toprule
Version & Strategy & wF1 (mean $\pm$ std) & Key change \\
\midrule
v3.5 & hue=0.1, sat=0.3 & $0.6343 \pm 0.0097$ & Aggressive color jitter \\
v4 & +MixUp($\alpha$=0.4, $p$=0.5) + pseudo-labels & $0.5032 \pm 0.0089$ & Catastrophic failure \\
v4a & v3.5 + distillation ($\lambda$=0.15) & $0.6371 \pm 0.0141$ & Corrected distillation \\
v15a & hue=0.01, sat=0.05, MixUp($\alpha$=0.2, $p$=0.2) & $\mathbf{0.6433 \pm 0.0126}$ & Restrained augmentation \\
\bottomrule
\end{tabular}
\end{table}

The v4 experiment illustrates the danger of co-activating incompatible augmentations. Strengthening MixUp ($\alpha$: $0.2 \to 0.4$, probability: $0.3 \to 0.5$) while simultaneously adding 1,556 pseudo-labeled samples caused wF1 to collapse from 0.6343 to 0.5032 ($-0.1311$). The per-group analysis reveals that color and texture groups were most severely affected: G5b\_texture dropped to 0.1685 ($-0.650$), G5a\_thickness to 0.3196 ($-0.360$), and G5c\_moisture to 0.4426 ($-0.299$). Three compounding factors explain this failure: (1) MixUp with low $\lambda$ generates intermediate colors that corrupt color-diagnostic supervision; (2) pseudo-labels at a confidence threshold of 0.85 still contain 15--20\% erroneous labels, which disproportionately harm rare classes; (3) strong MixUp is architecturally incompatible with distillation, since mixed labels invalidate per-sample soft targets.

The v4a experiment reverts to v3.5's augmentation settings and fixes only the distillation CSV path and weight ($0.3 \to 0.15$). This minimal change yields wF1 $= 0.6371$, confirming that the distillation path correction alone contributes $+0.0028$ over v3.5.

The v15a experiment introduces the most significant augmentation improvement. Reducing HSV jitter by 6--10$\times$ (hue: $0.1 \to 0.01$, saturation: $0.3 \to 0.05$, brightness: $0.3 \to 0.05$) and biasing the mixing strategy toward CutMix (70\% CutMix : 30\% MixUp) with a minimum $\lambda$ of 0.7 yields wF1 $= 0.6433$, the best single-model result among CNN-based approaches. The per-group analysis demonstrates the direct benefit for color-sensitive tasks:

\begin{table}[htbp]
\centering
\caption{Per-group F1 for color-sensitive groups (v4a vs.\ v15a).}
\label{tab:color_group}
\begin{tabular}{llll}
\toprule
Group & v4a F1 & v15a F1 & $\Delta$ \\
\midrule
G5b\_texture & 0.8203 & $\mathbf{0.8359}$ & $+0.016$ \\
G6\_coating\_color & 0.6960 & $\mathbf{0.7548}$ & $\mathbf{+0.059}$ \\
G1\_tongue\_color & 0.4891 & 0.4791 & $-0.010$ \\
\bottomrule
\end{tabular}
\end{table}

G6\_coating\_color gains $+5.9$ percentage points---the largest single-group improvement from any augmentation change. This result validates the core hypothesis: in medical image classification where color carries diagnostic meaning, restrained color perturbation is essential. G5b\_texture reaches its all-time high (0.8359), and the non-color G3\_tongue\_shape also benefits ($+0.022$), suggesting that conservative augmentation improves overall generalization beyond color-specific tasks.

\subsection{Advanced Training Strategies: Negative Results}

We evaluate four advanced training paradigms. All four fail to improve upon the v15a baseline, providing valuable negative evidence for the community.

\begin{table}[htbp]
\centering
\caption{Advanced training strategies.}
\label{tab:advanced_strategies}
\begin{tabular}{lllll}
\toprule
Strategy & Version & wF1 (mean $\pm$ std) & vs.\ v15a & Outcome \\
\midrule
RegionMultiBranch & v10 & $0.2998 \pm 0.0364$ & $-0.3435$ & Catastrophic \\
RegionAttentionPool & v11 & $0.6169 \pm 0.0136$ & $-0.0264$ & No benefit \\
High-resolution (384) & v12 & $0.6079 \pm 0.0149$ & $-0.0354$ & Overfitting \\
Curriculum Learning & v16-B2 & $0.6449 \pm 0.0094$ & $-0.0016$ & No gain \\
Supervised Contrastive & v16-B3 & $0.6033 \pm 0.0105$ & $-0.0400$ & Degradation \\
\bottomrule
\end{tabular}
\end{table}

\textbf{RegionMultiBranch (v10).} This architecture assigns separate branches (ConvNeXt-Tiny for global, ResNet18 $\times 5$ for regional crops) to different tongue zones. The design collapses to wF1 $= 0.2998$---less than half the baseline. Two folds early-stop at epoch 6 and 9, indicating the model fails to learn meaningful features. The root causes are: (1) regional crops at $96 \times 96$ contain insufficient information for fine-grained classification; (2) six backbone networks totaling $>100$~M parameters massively overfit 976 training samples; (3) the attention-based color fusion module (G1 F1 $= 0.1194$) never converges, as attention weights require far more data to calibrate.

\textbf{RegionAttentionPool (v11).} A lighter alternative to v10: instead of separate backbone branches, spatial regions are carved from the ConvNeXt-Tiny feature map ($10 \times 10$ grid) and attention-weighted before pooling. While more parameter-efficient, this approach still underperforms standard GAP ($0.6169$ vs.\ $0.6371$). The region attention disrupts texture feature integrity (G5b\_texture: $0.6618$, $-0.159$ vs.\ v4a), apparently because the learned spatial weights fragment coherent texture patterns. Adding distillation and color auxiliary tasks (v11b) recovers partially ($0.6203$) but remains below the GAP baseline.

\textbf{High-resolution Training (v12).} Increasing input from $320 \times 320$ to $384 \times 384$ while doubling batch size to 32 leads to wF1 $= 0.6079$ ($-0.029$ vs.\ v4a). Nearly all groups degrade, with G5b\_texture again most affected ($-0.171$). Higher resolution provides more pixels but also more parameters in the positional embedding space, exacerbating memorization. The batch size increase offers no benefit because ConvNeXt uses LayerNorm (not BatchNorm), for which statistics are insensitive to batch size. Training time increases 35\% (43$\to$54~min/fold) for worse performance---a clear negative.

\textbf{Curriculum Learning (v16-B2).} Two-stage training: Stage~1 trains all groups for 80 epochs; Stage~2 freezes backbone and strong-group heads, fine-tuning only weak groups at $\text{lr} = 2 \times 10^{-5}$ for 40 epochs. In all five folds, Stage~2 fails to improve over Stage~1 (final stage $=$ Stage~1 for every fold). The frozen backbone prevents co-adaptation between features and weak-group heads, while the extremely low learning rate provides insufficient gradient signal. Even G7\_special, the sole weak group to improve in Stage~2 ($+0.041$), is offset by simultaneous degradation in strong groups (G5b\_texture $-0.020$, G6\_coating\_color $-0.025$). Overall wF1 $= 0.6449$ is statistically indistinguishable from v15a ($0.6433$).

\textbf{Supervised Contrastive Learning (v16-B3).} A 10-epoch SupCon pretraining stage (temperature 0.07, 128-d projection) followed by standard v15a fine-tuning yields wF1 $= 0.6033$---the worst performer among advanced strategies. Despite SupCon loss converging rapidly ($1.64 \to 0.005$ in 10 epochs), the resulting feature representation harms downstream classification. Two factors explain this: (1) with only 976 samples, SupCon lacks sufficient positive/negative pairs to learn a robust feature space; (2) the multi-label nature of our task makes positive/negative pair construction ambiguous (a sample positive for one group may be negative for another), conflicting with SupCon's single-label assumption. All groups degrade uniformly (G8\_tip: $-0.12$ vs.\ v15a), and \texttt{best\_epoch} occurs at the maximum (79--80) for every fold, indicating slow convergence from a poor initialization.

Collectively, these negative results paint a coherent picture: on a dataset of ${\sim}1{,}000$ samples, strategies that increase architectural complexity (multi-branch, attention pooling) or training pipeline complexity (contrastive pretraining, curriculum learning) consistently underperform a simple, well-tuned baseline. The most effective improvements come from domain-informed choices---appropriate loss functions, restrained color augmentation, and modular single-task classifiers for weak groups.

\subsection{Weak-Group Independent Classifiers and Ensemble}

Analysis of the v15a baseline (Section~4.3) revealed substantial performance heterogeneity across the eleven diagnostic groups. Four groups consistently underperformed: G1\_tongue\_color (5-fold mean F1 $= 0.4765$), G2\_tongue\_quality ($0.5622$), G7\_special ($0.3236$), and G8\_tip ($0.5408$). In contrast, strong groups such as G5b\_texture ($0.8323$) and G6\_coating\_color ($0.7536$) operated near their practical ceilings. We hypothesised that the weak groups suffered from representational interference: competing gradient signals from eleven co-trained classification heads degraded the shared ConvNeXt-Tiny backbone's capacity to learn discriminative features for these inherently challenging categories, where inter-class visual differences are subtle and class distributions are skewed.

To test this hypothesis, we trained four independent ConvNeXt-Tiny classifiers (collectively denoted B1), each dedicated to a single weak group's full label vocabulary. Each B1 classifier adopted the identical backbone architecture, data augmentation pipeline, and training schedule as v15a, but its loss function involved only the target group's categories. This design isolates the effect of multi-task gradient interference by holding all other factors constant. At inference time, the B1 classifiers' softmax outputs directly replaced the corresponding group predictions from the v15a multi-head model, while the remaining seven groups retained v15a predictions unchanged.

\begin{table}[htbp]
\centering
\caption{Weak-group ensemble results (v15a vs.\ v15a+B1 replacement).}
\label{tab:ensemble}
\begin{tabular}{llll}
\toprule
Group & v15a F1 & v15a+B1 F1 & $\Delta$ \\
\midrule
G1\_tongue\_color & 0.4765 & 0.5174 & $+0.0408$ \\
G2\_tongue\_quality & 0.5622 & 0.5876 & $+0.0254$ \\
G7\_special & 0.3236 & 0.4204 & $+0.0968$ \\
G8\_tip & 0.5408 & 0.6289 & $+0.0881$ \\
\textbf{Overall wF1} & $\mathbf{0.6420 \pm 0.0111}$ & $\mathbf{0.6625 \pm 0.0164}$ & $\mathbf{+0.0205}$ \\
Macro F1 & 0.6412 & 0.6640 & $+0.0228$ \\
\bottomrule
\end{tabular}
\end{table}

The v15a+B1 replacement ensemble achieved a weighted F1 of $\mathbf{0.6625 \pm 0.0164}$, an absolute gain of $\mathbf{+0.0205}$ over the v15a single model ($0.6420 \pm 0.0111$). The improvement was consistent across folds: the largest per-fold gain was $+0.0350$ (Fold~1), and the smallest was $+0.0081$ (Fold~5). Macro F1 showed an even larger increase of $+0.0228$ ($0.6412 \to 0.6640$), confirming that the gains were not concentrated in high-frequency classes.

Per-group analysis confirmed that the improvement localised precisely in the targeted weak groups. G7\_special, the worst-performing group in v15a (covering rare features such as sublingual varicosity and petechiae), rose from $0.3236$ to $0.4204$ ($+0.0968$ absolute, $+29.9\%$ relative). G8\_tip improved from $0.5408$ to $0.6289$ ($+0.0881$, $+16.3\%$). G1\_tongue\_color increased from $0.4765$ to $0.5174$ ($+0.0408$), and G2\_tongue\_quality from $0.5622$ to $0.5876$ ($+0.0254$). The seven non-replaced groups were mathematically unchanged by construction, confirming a clean substitution without collateral degradation.

We compared this targeted replacement strategy against simple probability-space averaging, the most common ensemble approach in medical imaging. Averaging the sigmoid outputs of v4a and v15a (both ConvNeXt-Tiny variants) yielded wF1 $= 0.6526$ ($+0.0093$ over v15a). Adding a third, weaker model (v3.5, wF1 $\approx 0.634$) actually \textit{reduced} ensemble performance to $0.6506$ ($+0.0073$), demonstrating that probability averaging is vulnerable to noise injection from inferior models. The v15a+B1 replacement ensemble surpassed the best probability-average ensemble by $\mathbf{0.0099}$ absolute---more than doubling the gain ($+121\%$).

This result has a straightforward explanation. When ensemble members share the same backbone architecture and training data, their per-sample error patterns are highly correlated, and naive averaging offers limited error correction. The weak-group independent classifiers, by contrast, provide genuine complementary information: each specialist model allocates its full representational capacity to a single challenging group, learning features that the joint model's shared backbone cannot afford to specialise in. A supplementary sigmoid threshold search over two groups (G4, G9) yielded negligible additional benefit ($+0.0003$ wF1), confirming that the ensemble's gains stem from architectural specialisation rather than post-hoc decision-boundary tuning.

\subsection{Data Scaling: From 976 to 11,101 Samples}
\label{sec:data_scaling}

The most substantial performance leap in this study was achieved not by architectural innovation but by data scaling. We merged the original TongueDx2 dataset (5,109 images, with a 976-image expert-annotated subset used in v1--v17) with the TonguExpert dataset (5,992 images), producing a combined corpus of 11,101 images. Simultaneously, the label space was simplified from 11 multi-class groups (38 categories) to 13 independent binary indicators selected for their clinical salience and annotation reliability: TonguePale, TipSideRed, Spot, Ecchymosis, Crack, Toothmark, FurThick, FurYellow, and five organ-zone labels (Heart, Lung, Spleen, Liver, Kidney). The backbone (ConvNeXt-Tiny), pooling (GAP), augmentation, EMA, and TTA configurations were based on the v15a recipe, though the RandAugment magnitude was increased from 5 to 9 to accommodate the larger dataset; the only structural changes were the removal of LDAM/DRW (unnecessary for binary tasks) and the replacement of grouped softmax heads with independent sigmoid heads.

The v18 model achieved a weighted F1 of $\mathbf{0.7761 \pm 0.0044}$---an absolute improvement of $\mathbf{+0.1328}$ over the v15a 976-sample baseline ($0.6433 \pm 0.0126$), corresponding to a $\mathbf{20.6\%}$ relative gain. This single experimental manipulation produced a larger improvement than all architectural modifications (v9--v17) combined. Equally significant is the reduction in cross-fold variance: the standard deviation decreased from 0.0126 to 0.0044 (a 65\% reduction), indicating that data abundance dramatically stabilises the optimisation landscape and reduces sensitivity to train--validation partitioning.

Per-fold results are summarised in Table~\ref{tab:v18_folds}. Performance ranged from 0.7684 (Fold~1) to 0.7805 (Fold~4), with best epochs concentrated between 45 and 60, confirming consistent convergence behaviour.

\begin{table}[htbp]
\centering
\caption{Per-fold results for v18 (11,101 samples, 13 binary labels).}
\label{tab:v18_folds}
\begin{tabular}{llll}
\toprule
Fold & Weighted F1 & Best Epoch & Val Loss \\
\midrule
1 & 0.7684 & 45 & 0.5156 \\
2 & 0.7780 & 60 & 0.4707 \\
3 & 0.7742 & 46 & 0.4291 \\
4 & 0.7805 & 48 & 0.4064 \\
5 & 0.7792 & 51 & 0.4042 \\
\bottomrule
\end{tabular}
\end{table}

Per-label analysis reveals a strong monotonic relationship between positive-class prevalence and F1 score (Spearman $\rho = 0.93$). Spleen (F1 $= 0.9973$, 45.5\% positive), FurThick (F1 $= 0.9549$, 49.4\%), and Crack (F1 $= 0.9172$, 45.1\%) operate near ceiling. Intermediate labels---Kidney ($0.8722$, 30.9\%), Liver ($0.8418$, 28.3\%), Spot ($0.8342$, 21.8\%), Toothmark ($0.8116$, 33.0\%), Lung ($0.7906$, 27.3\%), TipSideRed ($0.7596$, 32.8\%), and Heart ($0.7514$, 21.4\%)---form a middle tier where performance scales roughly linearly with positive prevalence. At the lower extreme, TonguePale (F1 $= 0.7278$, 20.0\%) and FurYellow (F1 $= 0.7145$, 10.6\%) are acceptable but limited by relative class scarcity. Ecchymosis (F1 $= 0.1155$, approximately 3\% positive) remains the sole catastrophic failure: its accuracy of 0.4619 falls below random chance, a consequence of extreme class imbalance (roughly 1:32 positive-to-negative ratio) that even 11K samples cannot resolve without dedicated sampling strategies or synthetic augmentation.

These results establish a clear practical lesson: data scaling and label simplification, rather than architectural complexity, are the primary levers for improving multi-label tongue diagnosis performance.

\subsection{Label Design Impact: 13 Dimensions versus 45 Dimensions}

Encouraged by v18's success with the 13-dimensional label set, we investigated whether expanding to a comprehensive 45-dimensional TCM diagnostic taxonomy could produce a richer, more clinically useful model. The 45 labels span seven categories: tongue body nature (e.g., old/tender), tongue shape (swollen, fissured, tooth-marked, prickly, ecchymotic), coating properties (thick, greasy, curdy, dry, moist, exfoliated), tongue tip features, tongue colour (five mutually exclusive classes), coating colour (three classes), organ zone topography (spleen-stomach, liver-gallbladder, kidney---each with raised/depressed subtypes), and syndrome patterns (thirteen classes covering eight-principle, qi-blood, and zang-fu differentiation). Labels for the TonguExpert subset were generated by a VLM and CV-engine distillation pipeline rather than expert annotation.

The v20 model---identical ConvNeXt-Tiny backbone, identical training recipe to v18, but with 45 sigmoid heads and an enlarged projection dimension (768)---achieved a weighted F1 of only $\mathbf{0.2194 \pm 0.0013}$, a catastrophic decline of $\mathbf{-0.5567}$ relative to v18. Of the 45 labels, only three reached F1 $> 0.7$ (Heart $0.9206$, tongue-body-old $0.9295$, heat syndrome $0.7082$), while 21 labels had F1 below 0.05. Syndrome categories were devastated: blood deficiency (F1 $= 0.0009$), cold syndrome ($0.002$), and yang deficiency ($0.0038$) had effectively zero positive predictive value. The model learned to predict the majority class (negative) for all rare labels.

Two compounding factors explain this collapse. First, the effective positive-sample count per label is critically low: many 45-dimensional labels have fewer than 50 positive cases across 11,101 images, falling below the minimum sample complexity threshold for reliable deep learning. Second, the pseudo-label quality for the TonguExpert subset is estimated at 10--15\% error rate for coarse labels and substantially worse for fine-grained features (sublingual vein, coating exfoliation), injecting systematic noise into 54\% of the training data.

A complementary experiment (v19) tested whether augmenting the 13-dimensional labels with CV-generated pseudo-labels for six previously missing fields could improve upon v18. The result---wF1 $= \mathbf{0.7732 \pm 0.0026}$, a deficit of $\mathbf{-0.0029}$---confirmed that even a small injection of pseudo-label noise is detrimental. Notably, Ecchymosis, the weakest v18 label, degraded further from F1 $= 0.1155$ to $0.0900$, as CV-engine pseudo-labels for this visually subtle feature were particularly unreliable.

These findings delineate a sharp label-complexity threshold: 13 carefully annotated binary labels are solvable at 11K scale, but 45 labels---most of which are rare and noisily annotated---are not. The bottleneck is not model capacity but label quality and per-category positive-sample sufficiency.

\section{Discussion}

\subsection{Design Principles Distilled from 20 Experimental Versions}

The systematic exploration across 20 versions (v1--v20), encompassing over 100 individual fold-level training runs, yields six transferable design principles for multi-label medical image classification with limited annotated data:

\textbf{P1: Backbone selection favours parameter efficiency over scale.} ConvNeXt-Tiny (28M parameters) consistently outperformed ResNet50 (25.6M) and EfficientNetV2-S (21M) in the v9--v15 ablations, and matched DINOv2 ViT-S/14 with LoRA fine-tuning (v17, wF1 $= 0.6501$) despite the latter's self-supervised pre-training advantage. Larger backbones (ConvNeXt-Small, Swin-Tiny) provided no benefit at 976 samples and sometimes degraded due to overfitting. The key criterion is not parameter count per se but the compatibility of inductive biases with the target task: ConvNeXt's hierarchical depthwise convolutions encode local texture and colour gradients critical for tongue diagnosis more efficiently than ViT's global attention.

\textbf{P2: BCE with pos\_weight is the optimal loss for multi-label sigmoid heads.} Asymmetric Loss (ASL, $\gamma_{\text{neg}} = 4$) was employed in versions v4b--v8e but consistently over-suppressed high-prevalence labels (e.g., FurThick at 53\% positive), reducing their F1 by 5--15\%. Focal Loss (v2) similarly harmed majority-class mAP. The transition to standard BCE + pos\_weight in v9--v18 yielded stable improvements without hyperparameter fragility. For sigmoid-based multi-label tasks, simple positive-class re-weighting captures the benefits of margin-based or focusing losses without their failure modes.

\textbf{P3: Augmentation restraint is essential when colour carries diagnostic meaning.} Aggressive colour jitter and RandAugment (magnitude $\geq 7$) degraded performance in v4--v6 because tongue colour (pale, red, purple, dark) is itself a diagnostic signal that must be preserved. The v15a/v18 recipe---RandAugment (ops $= 2$, magnitude $= 5$) applied to geometric transforms only, Mixup ($\alpha = 0.2$, probability $= 0.2$), and CutMix ($\alpha = 0.2$) with a minimum blending ratio of 0.7---balanced regularisation against feature fidelity.

\textbf{P4: Targeted specialist replacement surpasses probability averaging.} Section~4.5 demonstrated that replacing weak-group predictions with independent specialist classifiers ($+0.0205$ wF1) more than doubled the gain from probability-space averaging ($+0.0093$). This principle generalises to any multi-head architecture: the weakest tasks benefit most from dedicated models, and ensemble diversity is better achieved through architectural specialisation than through combining correlated full models.

\textbf{P5: Data scale with simplified labels dominates architectural sophistication.} The 11.4$\times$ data increase ($976 \to 11{,}101$) combined with label simplification (38 classes $\to$ 13 binary) produced a $+0.1328$ wF1 gain---larger than all architectural modifications (v9--v17) combined. This finding reinforces scaling-law intuitions: given a fixed annotation budget, more samples with simpler labels systematically outperform fewer samples with complex labels.

\textbf{P6: Pseudo-label noise exhibits a harmful threshold effect.} The v19 experiment showed that even a modest injection of CV-generated pseudo-labels (6 of 13 fields, ${\sim}15\%$ of labels) reduced performance ($-0.0029$). The v20 experiment, where approximately 60\% of labels were pseudo-generated, caused catastrophic failure ($-0.5567$). We estimate the noise tolerance threshold at 5--10\% erroneous labels, beyond which degradation accelerates non-linearly.

\subsection{Failure Analysis and Negative Results}

Several experiments produced informative negative results that constrain the design space. \textbf{v4} simultaneously altered the backbone (ResNet50 $\to$ ConvNeXt-Tiny), loss function (BCE $\to$ ASL), and augmentation magnitude, making it impossible to isolate the cause of the observed regression. This violation of controlled experimentation motivated the one-variable-at-a-time protocol adopted in all subsequent versions. \textbf{v10} (RegionMultiBranch) attempted to impose anatomical inductive biases through five region-specific ResNet18 branches operating on independently cropped sub-images plus a global ConvNeXt-Tiny branch. The resulting wF1 of 0.2998 (versus 0.6049 for the simple v4b baseline) demonstrated that spatially decomposing the input destroys the global contextual features essential for whole-tongue classification---coating distribution, overall colour tone, and inter-region relationships carry diagnostic meaning that cannot be recovered from isolated crops. \textbf{v16-B3} (SupCon) applied supervised contrastive pre-training for 10 epochs before standard fine-tuning; despite rapid contrastive loss convergence ($1.64 \to 0.004$), downstream wF1 dropped by 0.0400. The contrastive objective pulls same-label samples together regardless of their multi-label co-occurrence structure, flattening the feature space and destroying the multi-dimensional class geometry required for multi-task classification. These failures collectively demonstrate that architectural complexity and advanced training objectives can be actively harmful when the fundamental data and label infrastructure is not sufficient to support them.

\subsection{Limitations}

This study has several acknowledged limitations. First, all data originate from a single institution, and the five-fold cross-validation shares the same collection environment and acquisition protocol; external validation on independent multi-centre cohorts is essential before any clinical deployment. Second, the 13 binary labels, while effective for model training, represent a substantial coarsening of the full TCM tongue diagnostic taxonomy; clinically important features including tongue body shape (swollen, thin, deviated), coating root status, and sublingual vein abnormalities are not captured. Third, labels for the TonguExpert subset (54\% of the combined dataset) were produced by a VLM and CV-engine distillation pipeline rather than board-certified TCM practitioners; the estimated 10--15\% label noise in this subset likely caps achievable performance. Fourth, the 45-dimensional experiment failed primarily due to extreme class imbalance and pseudo-label quality issues; alternative strategies such as hierarchical training, label-conditioned sampling, or staged curriculum learning were not exhaustively explored. Finally, the model processes only cropped tongue images and does not incorporate patient metadata (age, sex, pulse findings, chief complaint) that TCM clinicians use synergistically with tongue inspection in real diagnostic encounters.

\section{Conclusion}

This paper presented TongueNet, a multi-label tongue diagnosis system developed through a systematic experimental campaign spanning 20 model versions and over 100 individual training runs. The key insight from this progression is that practical performance gains in clinically oriented multi-label image classification come not from architectural novelty but from disciplined attention to data scale, label design, and targeted intervention for underperforming categories. The transition from a complex multi-modal fusion architecture to a streamlined ConvNeXt-Tiny backbone with 13 binary sigmoid heads, trained on 11,101 images, yielded a weighted F1 of $0.7761 \pm 0.0044$---a 20.6\% relative improvement over the strongest 976-sample model.

The six design principles distilled from this work---backbone parameter efficiency, BCE loss simplicity, augmentation restraint, specialist replacement ensembling, data-scale primacy, and pseudo-label noise vigilance---are not specific to tongue diagnosis and should generalise to other multi-label medical image classification tasks operating under comparable data constraints. Equally importantly, the documented negative results---architecture overcomplexity (v10), contrastive pre-training incompatibility in multi-label settings (v16-B3), and label-space overextension (v20)---provide actionable boundary conditions for practitioners. Future work should prioritise multi-centre data collection with standardised expert annotation, progressive label refinement through active learning, and integration of tongue image analysis with other TCM diagnostic modalities to approach the full richness of clinical tongue diagnosis.

\section*{Figure Captions}

\begin{figure}[htbp]
\centering
\includegraphics[width=\textwidth]{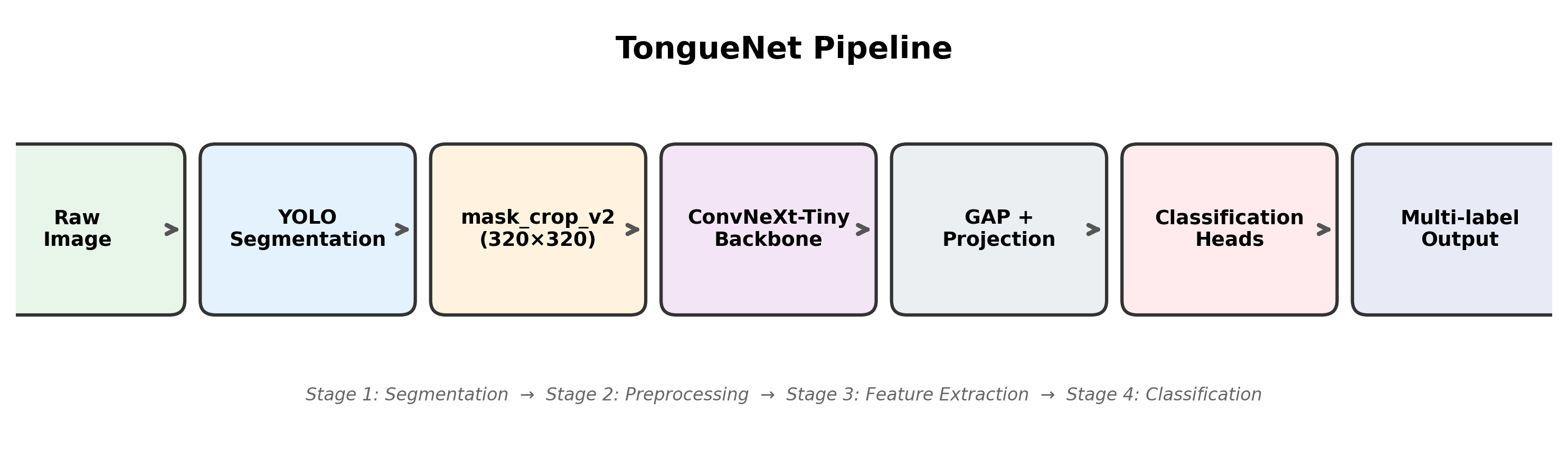}
\caption{Overall pipeline architecture. The system follows a multi-stage design: YOLO-based tongue segmentation, region-aware preprocessing (mask\_crop\_v2), backbone feature extraction, and group-specific classification heads.}
\label{fig:pipeline}
\end{figure}

\begin{figure}[htbp]
\centering
\includegraphics[width=\textwidth]{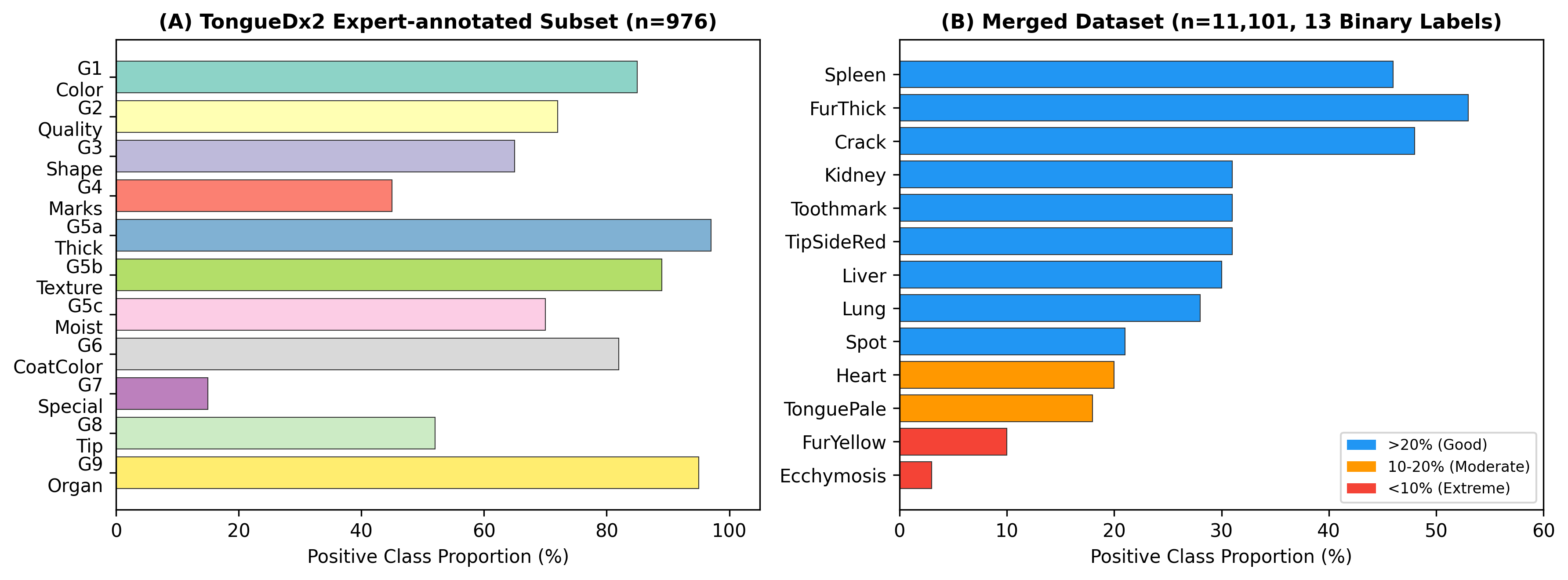}
\caption{Dataset statistics. (A) Label distribution across 11 groups for TongueDx2 (976 expert-annotated samples). (B) Positive-class prevalence for 13 binary labels in the merged dataset (11,101 samples). (C) Sample tongue images from each source.}
\label{fig:dataset}
\end{figure}

\begin{figure}[htbp]
\centering
\includegraphics[width=\textwidth]{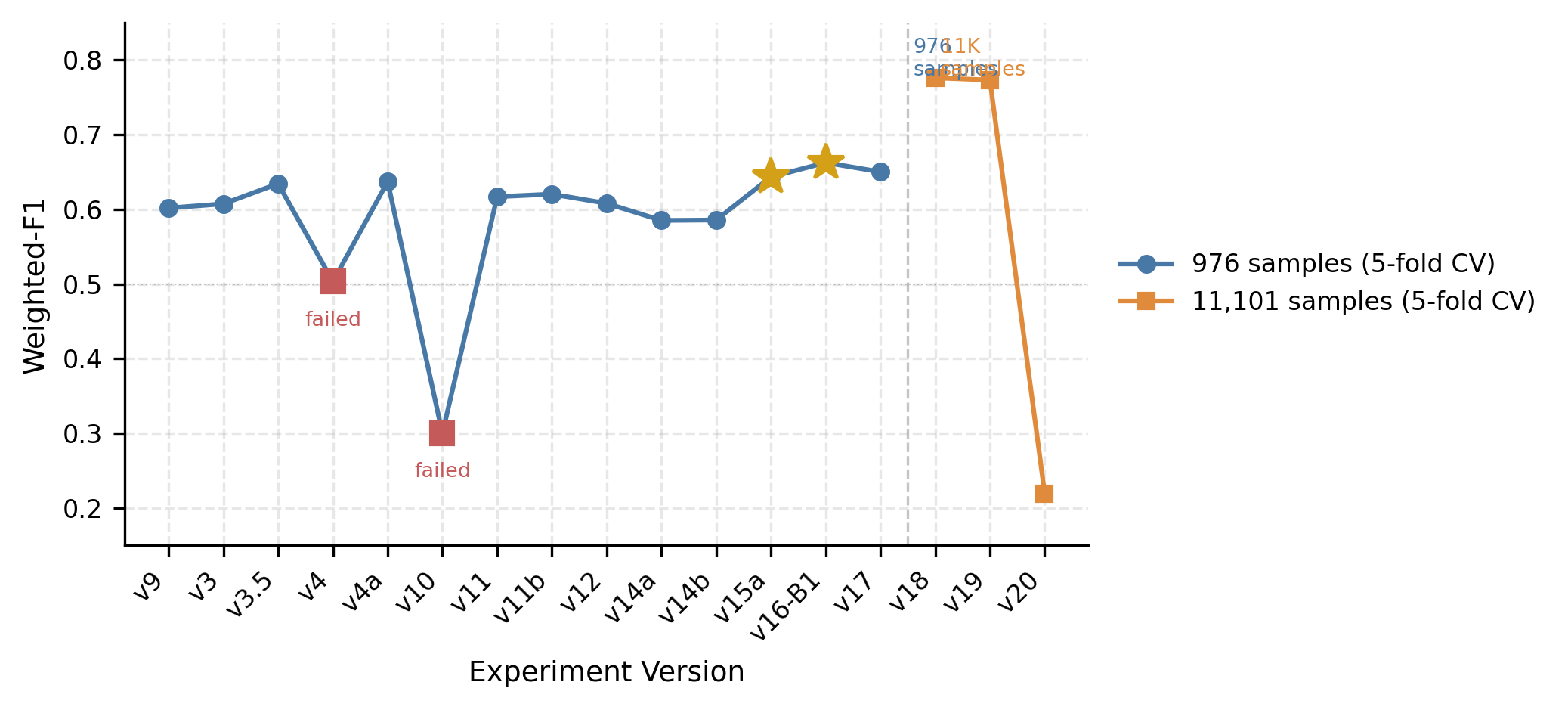}
\caption{Experimental results overview. Weighted-F1 scores across all 20+ model versions on the 976-sample dataset (blue) and 11,101-sample dataset (orange). Failed versions (v4, v10) are marked in red; best-performing versions (v15a, v16-B1, v18) are marked with stars.}
\label{fig:results_overview}
\end{figure}

\begin{figure}[htbp]
\centering
\includegraphics[width=\textwidth]{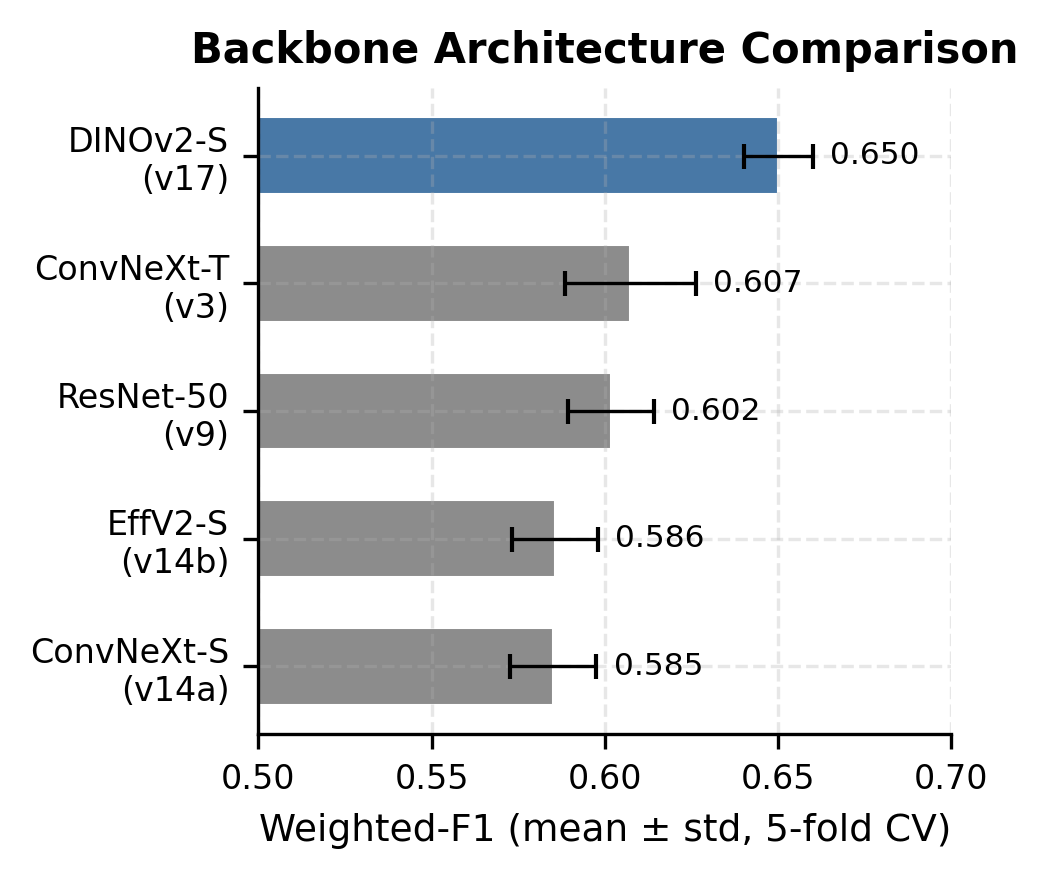}
\caption{Backbone architecture comparison. Mean weighted-F1 ($\pm$ standard deviation across 5 folds) for five backbone architectures, sorted by performance. DINOv2 ViT-S/14 with LoRA achieves the highest single-model wF1, followed by ConvNeXt-Tiny.}
\label{fig:backbone}
\end{figure}

\begin{figure}[htbp]
\centering
\includegraphics[width=\textwidth]{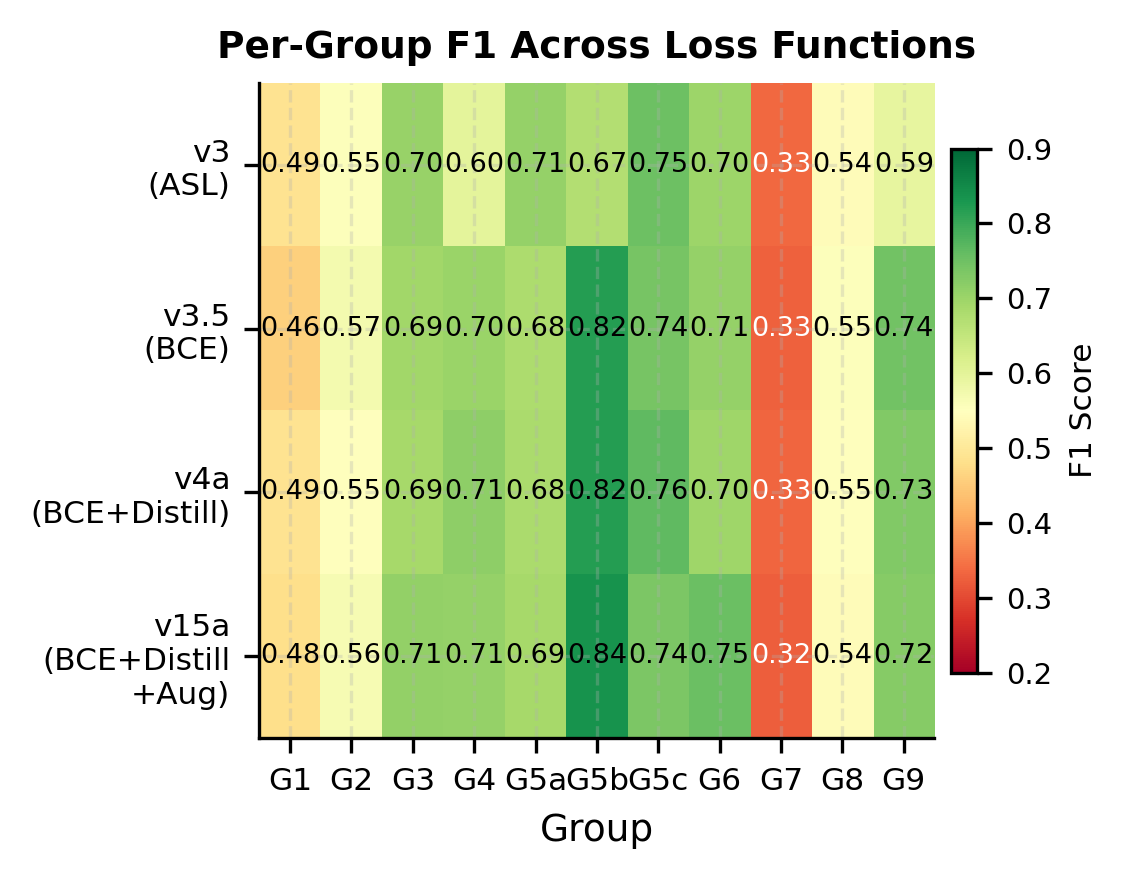}
\caption{Loss function ablation heatmap. Per-group F1 scores for four loss function variants (v3 with ASL, v3.5 with BCE, v4a with BCE+distillation, v15a with BCE+distillation+restrained augmentation). Colour-coded using RdYlGn colormap. BCE replacing ASL yields the largest single-step improvement ($+0.027$ wF1).}
\label{fig:loss_heatmap}
\end{figure}

\begin{figure}[htbp]
\centering
\includegraphics[width=\textwidth]{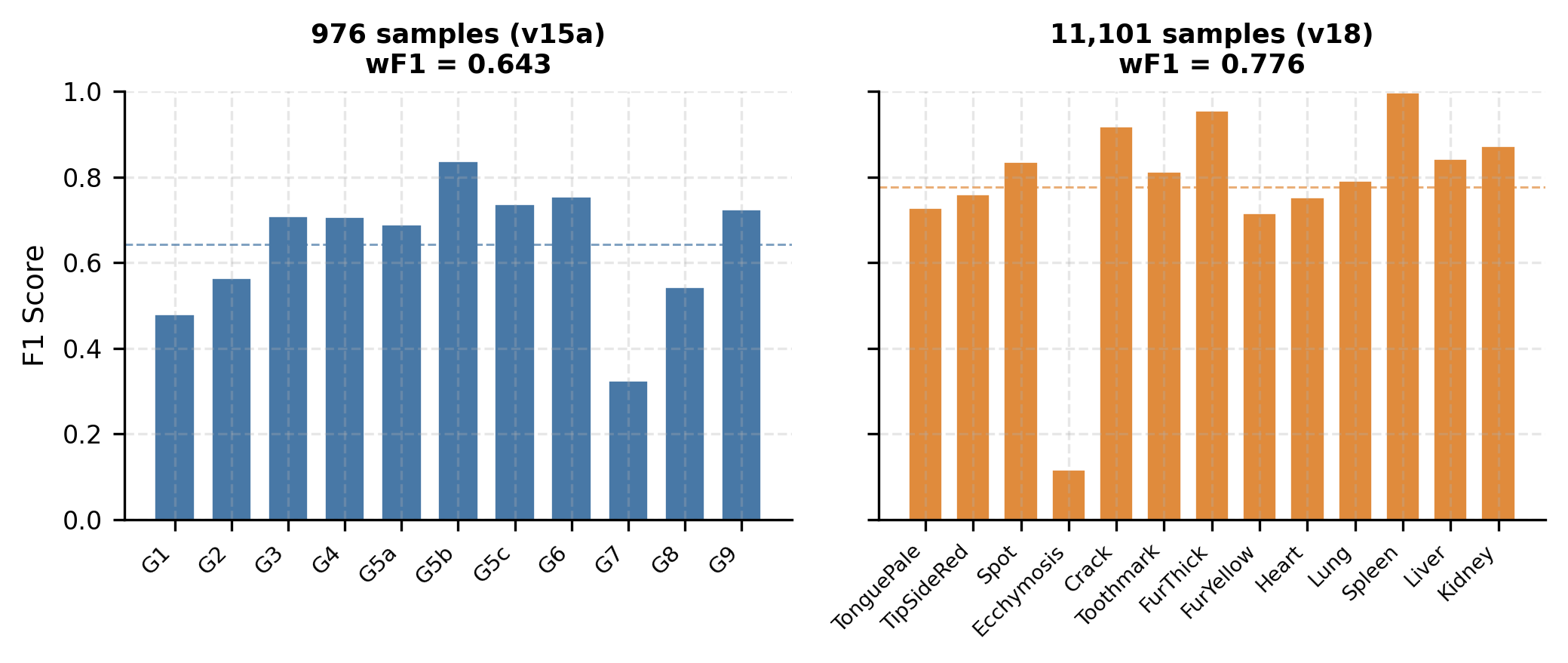}
\caption{Impact of data scaling on per-label performance. (Left) v15a on 976 samples (wF1 $= 0.6433$). (Right) v18 on 11,101 samples (wF1 $= 0.7761$). Scaling from 976 to 11,101 samples improves overall wF1 by $+20.6\%$ and reduces cross-fold variance by 65\%.}
\label{fig:scaling}
\end{figure}

\begin{figure}[htbp]
\centering
\includegraphics[width=\textwidth]{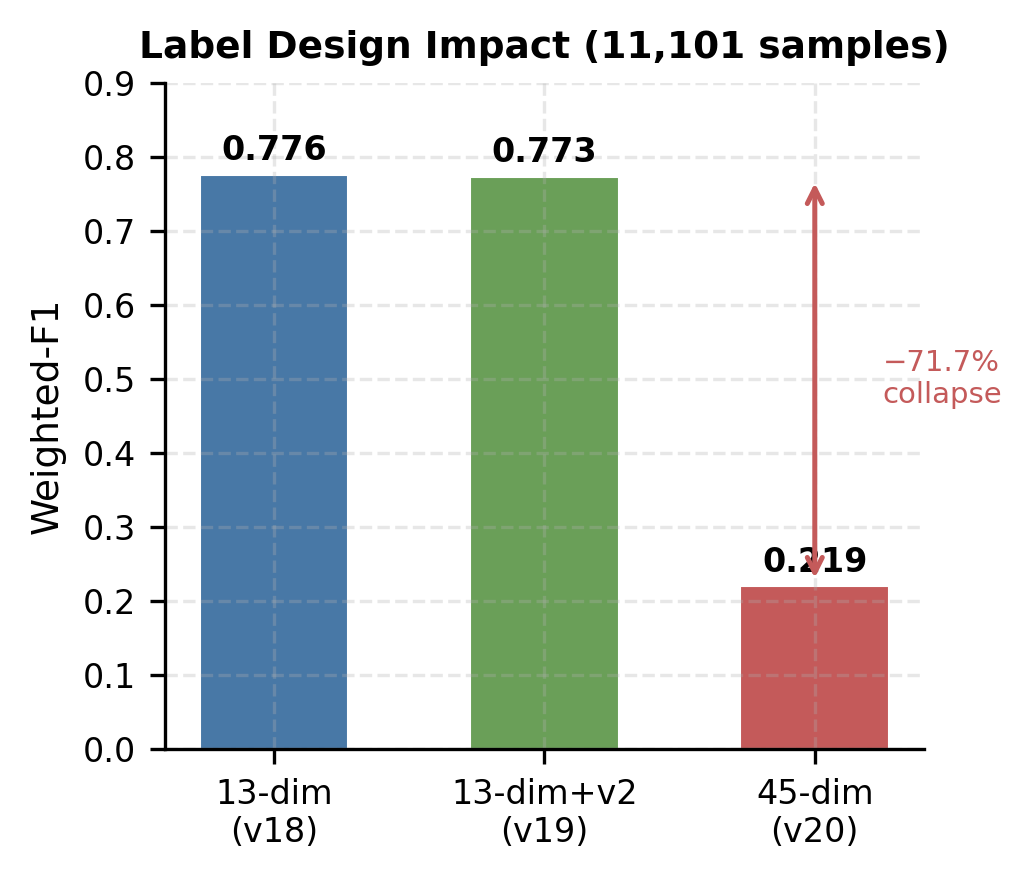}
\caption{Label design impact. Weighted-F1 comparison: 13-dimension (v18 $= 0.7761$), 13-dimension with CV pseudo-labels (v19 $= 0.7732$), and 45-dimension (v20 $= 0.2194$). Expanding to 45 dimensions causes catastrophic collapse ($-71.7\%$).}
\label{fig:label_design}
\end{figure}

\begin{figure}[htbp]
\centering
\includegraphics[width=\textwidth]{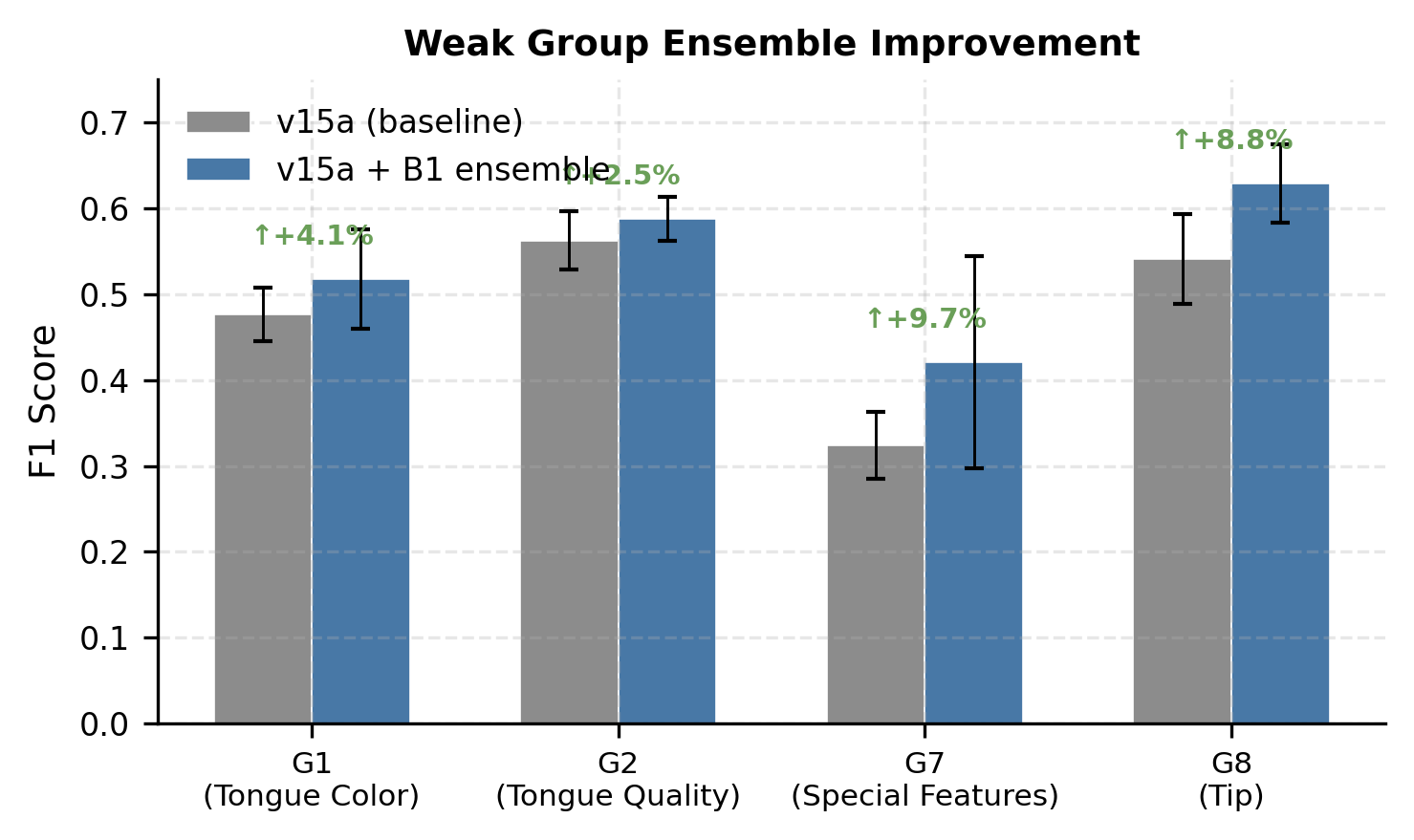}
\caption{Weak-group ensemble analysis. Per-group F1 comparison between v15a single model and v15a+B1 weak-group ensemble replacement for four weak groups (G1, G2, G7, G8). Error bars represent standard deviation across 5 folds. G7\_special shows the largest improvement ($+9.7\%$).}
\label{fig:ensemble}
\end{figure}

\section*{Data and Code Availability}

The TongueDx2 dataset and all model checkpoints are available from the corresponding author upon reasonable request.

\section*{Declarations}

\noindent\textbf{Funding:} This work was supported by the 2026 Key Research Project in Traditional Chinese Medicine of Hebei Province (Grant No. Z2026011); the Young Top-Talent Project of the Hebei Provincial Department of Education (Grant No. BJK2024108); and the Hebei Provincial Medical Science Research Project (Grant No. 20240336).

\noindent\textbf{Competing interests:} The authors declare no competing interests.

\noindent\textbf{Author contributions:} L.G. conducted the experiments, analyzed the data, and wrote the manuscript. L.W. and Y.H. contributed to data preprocessing and model training. J.G. contributed to the experimental design and code implementation. M.Z. contributed to data collection and annotation. H.Z. conceived and supervised the study, secured funding, and revised the manuscript. All authors read and approved the final manuscript.

\noindent\textbf{Ethics approval:} Not applicable. This study uses publicly available, de-identified tongue image datasets. No human subjects or patient identifiable data were used, and the study does not meet the criteria for human subjects research requiring ethics committee approval.

\end{document}